\documentclass[letterpaper, 10 pt, conference]{ieeeconf}  
\IEEEoverridecommandlockouts                              
\usepackage{graphicx}
\usepackage{epsfig} 
\usepackage{mathptmx} 
\usepackage{times} 
\usepackage{amsmath} 
\usepackage{amssymb}  

\usepackage{cite}
\usepackage{url} 
\usepackage{bm} 
\usepackage{booktabs} 
\usepackage{multirow}   
\usepackage{placeins}
\usepackage{enumerate}
\let\labelindent\relax 
\usepackage{enumitem}

\title{\LARGE \bf
Human-Centric Grasp State Assessment: \\ Toward Transferring Subjective Evaluation to Robots
}

\author{
Ryohei Kobayashi$^{1}$,
Kosei Isomoto$^{1}$,
Yuga Yano$^{1}$,
Yuichiro Tanaka$^{1,\ 2}$,
Hakaru Tamukoh$^{1,\ 2}$
    \thanks{$^{1}$ Ryohei Kobayashi, Kosei Isomoto, Yuga Yano, Yuichiro Tanaka and Hakaru Tamukoh are with Graduate School of Life Science and Systems Engineering, Kyushu Institute of Technology, 2-4 Hibikino, Wakamatsu, Kitakyushu, Fukuoka, 808-0196, Japan
        {\tt\small{ \{kobayashi.ryohei621, isomoto.kosei778, yano.yuuga158\}@mail.kyutech.jp, \{tanaka-yuichiro, tamukoh\}@brain.kyutech.ac.jp }}
    }%
    \thanks{$^{2}$ Yuichiro Tanaka and Hakaru Tamukoh are also affiliated with the Research Center for Neuromorphic AI Hardware,  Kyushu Institute of Technology, 2-4 Hibikino, Wakamatsu, Kitakyushu, Fukuoka, 808-0196, Japan}%
}

\usepackage{capt-of}
\makeatletter
\let\@oldmaketitle\@maketitle
\renewcommand{\@maketitle}{\@oldmaketitle
  \begin{center}
    \vspace{0.3em} 
    \label{fig:bg}
    \includegraphics[width=1.0\linewidth]{figs/icdl_introduction_v3.pdf}
    \captionof{figure}{
    Human-Centric Grasp State Assessment to bridge the gap between robot behavior and human expectations.
    }
    \label{fig:placeholder}
  \end{center}
  \vspace{-1.0em} 
}
\makeatother

\begin{document}
\maketitle
\thispagestyle{empty}
\pagestyle{empty}


\setcounter{figure}{1} 
\begin{abstract}
We propose a framework that transfers tacit human subjective criteria to robotic systems for the appropriate grasping of deformable objects. Achieving such behavior is challenging because a semantic gap exists between qualitative human expectations and quantitative robotic measurements. Conventional deep learning approaches for bridging this gap also require prohibitive amounts of manually annotated data for each newly encountered object. To address these challenges, our framework integrates a Vision-Language Model (VLM)-based semi-automated supervisor generator with a lightweight grasp state predictor, using a minimal set of human-annotated trials as contextual anchors to propagate subjective criteria to unannotated data. The prediction model then enables rapid online adaptation by sequentially estimating the grasp state from time-series tactile and grasping force measurements.
Through experiments on three representative deformable objects and a human evaluation study with 25 participants, we demonstrate the feasibility of the proposed framework for adjusting grasping force according to human-perceived grasp appropriateness in the evaluated task setting.


\end{abstract}

\section{INTRODUCTION}
In unstructured environments such as homes and convenience stores, the demand for mobile manipulators capable of handling daily necessities and foods has rapidly increased in recent years \cite{tanaka2020,tokuno2024,yano2024}. 
Such environments contain a wide variety of objects with different shapes, 
materials, and stiffness, many of which are not known in advance to robots. 
Consequently, autonomous robots must continually handle newly encountered objects. 

Since the physical properties of such objects, including stiffness and fragility, 
are unknown beforehand, determining an appropriate grasping strategy becomes a 
critical challenge. In particular, robots must regulate grasping forces to ensure 
stable interaction under uncertainty, avoiding both sliding and structural failure.
Grasping deformable objects requires a delicate balance because insufficient force causes sliding and excessive force causes structural damage \cite{zhu2025}. 

Beyond ensuring such basic physical safety, robots must also align with tacit human subjective criteria, as mere numerical thresholds often fail to capture the nuanced expectations regarding preserved appearance and tactile integrity of an object.
This study focuses on grasping deformable objects as a representative task for realizing human-centric robotic manipulation.
These criteria vary inherently with the material and stiffness of an object.
For instance, a force that safely secures a firm object might crush a softer one, whereas a gentler force might provide insufficient friction and cause sliding. Because robots rely solely on objective and numerical internal measurements such as tactile distributions and torque, they lack a mechanism to directly map these human visual expectations into specific control parameters. This discrepancy creates a significant semantic gap between qualitative human criteria and quantitative robotic measurement, as illustrated in the Issue of Fig. 1. Such misalignment between robot behavior and human expectations can reduce user trust in human-robot interaction \cite{hancock2011}.


A straightforward approach to bridging this gap is to manually define physical 
thresholds for each object. However, this approach is fundamentally unscalable 
in open environments with diverse objects, and human subjective evaluation cannot 
be reduced to simple numerical boundaries. Cui et al. addressed this problem by 
abstracting grasp states into discrete categories--sliding, appropriate, and 
excessive--using visual and tactile information \cite{cui2020}. However, their 
deep learning approach requires large-scale annotated data, making it impractical 
to adapt to newly encountered objects.

To enable continuous operation in such environments, a human-centric approach requires a novel learning foundation. Going beyond merely evaluating task success or failure, this foundation must dynamically bridge the semantic gap, enabling robots to rapidly acquire and align with human perception for newly encountered objects. 
Consequently, this study proposes a framework that transfers human subjective evaluation criteria to robots with reduced annotation and retraining costs, thereby enabling grasping behavior that is better aligned with human-perceived appropriateness in the evaluated task setting.


The main contributions of this study are as follows:
\begin{itemize}
    \item We propose a method to reduce the annotation cost for newly encountered objects by implementing a semi-automated training data generation mechanism. This mechanism utilizes a Vision-Language Model (VLM) initialized with a small number of human subjective annotations.
    \item We realize a sequential framework for object-wise adaptation by integrating a lightweight time-series learning model.
\end{itemize}


\section{RELATED WORK}

This section reviews previous studies on the grasping of deformable objects with force and tactile feedback, the assessment of grasp state using visual-tactile fusion, and the generation of training data using VLM. 

\subsection{Subjective Criteria and Semantic Gap in Grasping}
While gripper force control and tactile feedback have been widely used to stabilize deformable object manipulation~\cite{narita2020,hogan2020,kaboli2016}, these approaches primarily address physical safety, such as preventing slip or reducing excessive grasping force, and do not fully capture human subjective evaluation criteria. In particular, even if slip is avoided, a grasp may still be perceived as inappropriate when visual deformation or loss of appearance exceeds human tolerance. 
Moreover, such human judgments depend on object-specific properties such as material and stiffness, and cannot be directly inferred from sensor values alone without human supervision~\cite{patni2024}. Our framework addresses this by correlating human subjective grasp states with corresponding tactile and grasping force measurements. In this sense, slip-based control and the proposed framework are complementary: the former maintains physical grasp stability, whereas the latter estimates human-perceived deformation tolerance for grasp-force adjustment.


\subsection{Efficiency and Adaptation in Grasp State Assessment}
Cui et al. proposed a 3D convolution-based visual-tactile fusion deep neural network (C3D-VTFN) that takes visual and tactile information as inputs to classify the grasp state into three discrete categories of sliding, appropriate, and excessive \cite{cui2020}. The C3D-VTFN processes visual and tactile time-series data using independent 3D CNNs and integrates the extracted spatiotemporal features through fully-connected layers. Notably, this method enables time-series processing for tactile data similar to visual data by treating the 3-axis tactile distribution as images. Furthermore, it learns effective fusion features from asynchronous heterogeneous data by absorbing the difference in sampling rates between the image and tactile sensors through the ratio of input frames. They constructed a large-scale Grasp State Assessment (GSA) dataset consisting of 20,000 samples across 16 objects, achieving a classification accuracy of 99.98\%.

On the other hand, deep neural network-based methods assume large-scale prior dataset collection and require substantial time for model training. 
Therefore, instant online adaptation remains difficult in real-world environments where newly encountered objects are constantly introduced. 

To address these limitations, our framework correlates human subjective evaluation with tactile and gripper force measurements during training. This strategy enables a lightweight model to perform inference using only these mechanical data, which minimizes data requirements and achieves rapid adaptation to newly encountered objects.

\FloatBarrier
\begin{figure*}[!t]
  \centering
  \includegraphics[width=2.0\columnwidth]{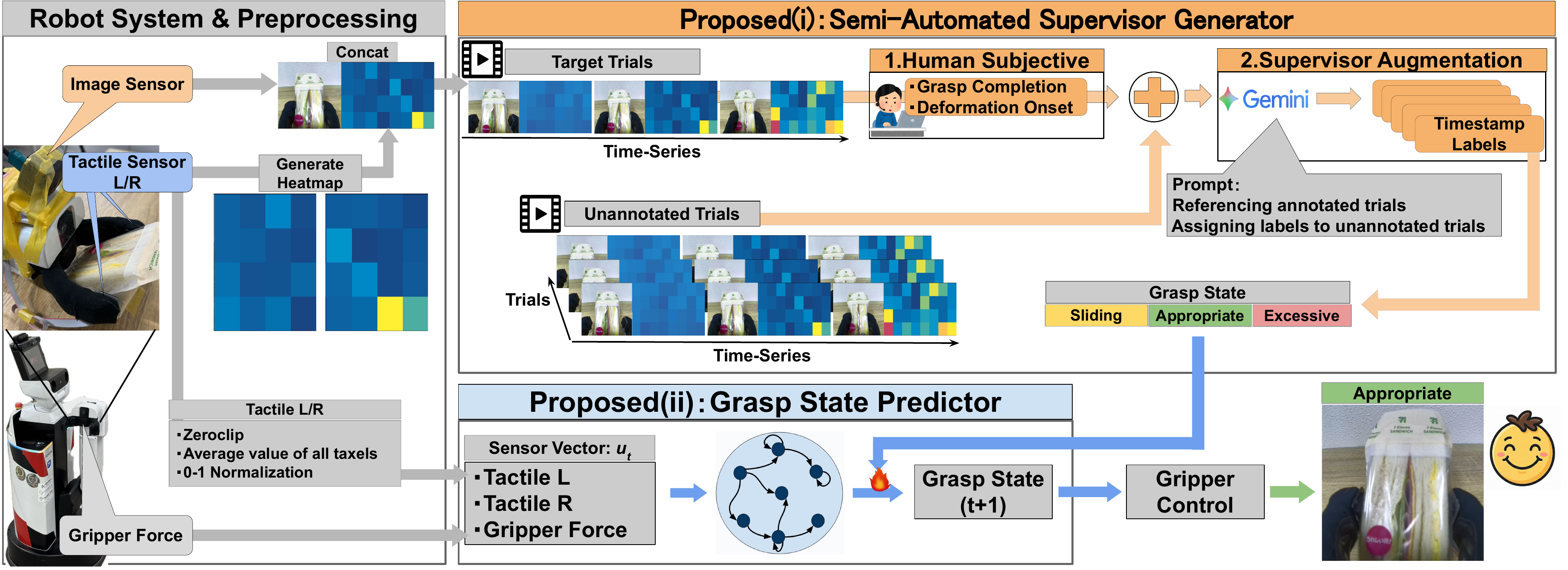}
  \caption{Proposed framework.}
  \label{fig:propose_method}
\end{figure*}

\subsection{Supervisor Generation using VLMs} \label{sec:vlm}

To resolve the data collection bottleneck identified in deep learning approaches, our framework focuses on semi-automated supervisor data generation.
VLMs demonstrate high performance in image and video understanding; thus, researchers have recently utilized them to generate supervisor data \cite{xu2025, silva2026}. Based on linguistic instructions, VLMs can evaluate ambiguous events while considering context and situations, and they can even generate explanations and reasoning for their judgments based on visual inputs.

However, previous studies report that VLM outputs often suffer from unfounded hallucinations and wavering judgments in ambiguous situations \cite{rohrbach2018}. 
The zero-shot inference performance of these models regarding intricate physical concepts often fails to align with human expectations \cite{gao2023}. 
To mitigate these issues, Silva-Rodríguez et al. proposed a semi-supervised approach that utilizes pre-trained text knowledge as an anchor, ensuring that class representations do not deviate significantly from prior knowledge during low-cost, few-shot adaptation \cite{silva2026}. Xu et al. emphasized the importance of structuring the output format, such as forcibly mapping responses to predefined categories, to ensure label reliability within automated pipelines \cite{xu2025}.

Our framework ensures supervisor generation by integrating structured outputs with a human subjective anchoring strategy. This approach enables stable, human-aligned training data generation for newly encountered objects with minimal manual intervention.


\section{PROPOSED FRAMEWORK}
Figure \ref{fig:propose_method} illustrates the proposed framework. To support object-wise adaptation with limited annotation, the framework integrates two main modules: 
\begin{itemize}
    \item Supervisor Generator (SG) that generates training data starting from a limited number of human subjective annotations.
    \item Grasp State Predictor (GSP) that predicts one-step-ahead the grasp state of sliding, appropriate, and excessive from time-series tactile and grasping force data. 
\end{itemize}
Rather than replacing human annotation, the SG reduces labeling costs. Furthermore, employing a lightweight model for the GSP keeps the computational cost of online adaptation practical for continuous operation.

\subsection{Dataset Preparation} \label{sec:dataset_preparation}

In the proposed framework, the robot measures the grasp state using visual, tactile, and grasping force information. As shown in the Robot System \& Preprocessing of Fig. \ref{fig:propose_method}, the data streams from the camera, tactile sensors, and gripper force, along with the annotated grasp states (sliding, appropriate, and excessive), are formatted into structures suitable for the SG and GSP.
Specifically, the dataset consists of RGB images from a hand-mounted camera, the z-axis displacement values indicating the pushing direction from the left and right tactile sensors, and the applied grasping force.
The recorded RGB images are compiled into videos. 

To establish the ground truth (GT) labels for the dataset, a human annotator observes these RGB videos and manually assigns two specific timestamps, namely the grasp completion time $t_{grasp}$ and the deformation onset time $t_{deform}$.
These timestamps are defined as follows:
\begin{itemize}
    \item $t_{grasp}$: The state where the gripper securely holds the object. The moment one can visually judge that no gap exists between the gripper and the object.
    \item $t_{deform}$: The moment one can judge that visual deformations, such as dents on the object outline or surface, begin to appear due to increasing grasping force, indicating a loss of its original appearance.
\end{itemize}

\begin{figure}[t]
  \centering
  \includegraphics[width=0.8\columnwidth]{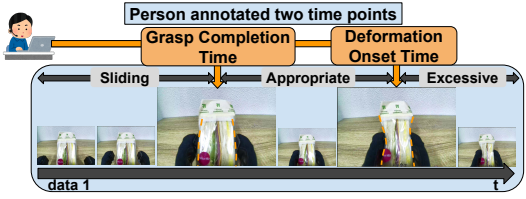}
  \caption{Relationship between the grasp state and timestamp labels.}
  \label{fig:annotation}
\end{figure}

As shown in Fig. \ref{fig:annotation}, these timestamps correspond to the three discrete metrics proposed by Cui et al. \cite{cui2020} and are expressed as follows:
\begin{itemize}
    \item Sliding: $t < t_{grasp}$
    \item Appropriate: $t_{grasp} \le t < t_{deform}$
    \item Excessive: $t \ge t_{deform}$
\end{itemize}

\begin{figure}[t]
  \centering
  \includegraphics[width=0.7\columnwidth]{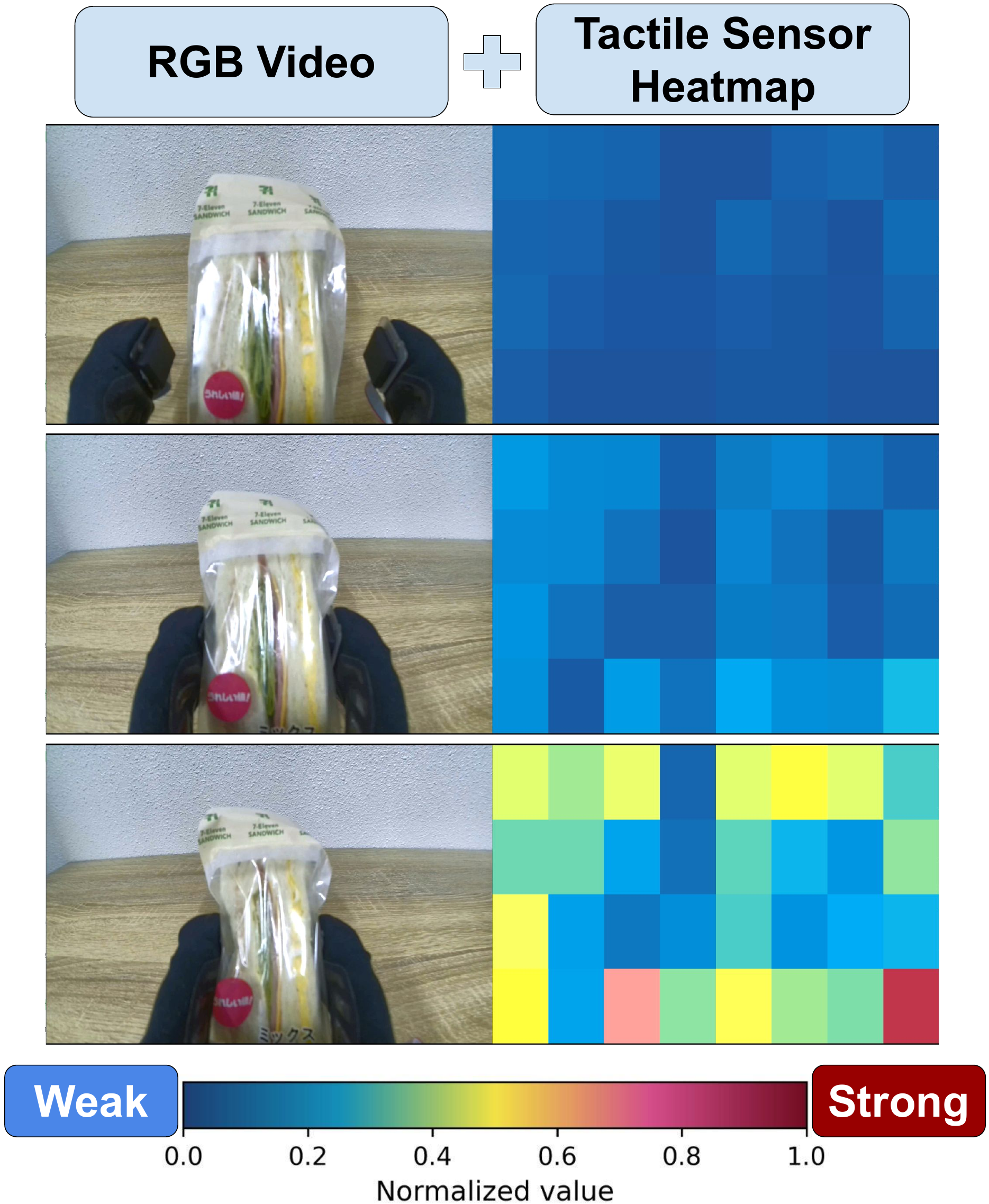}
  \caption{Concatenated video of the hand camera view and the tactile heatmap.}
  \label{fig:video_input}
\end{figure}

Two distinct data formats are constructed from the acquired multimodal data for training.
The first format is video data designed for the SG to capture spatiotemporal state transitions. RGB images from the hand camera are horizontally concatenated with a heatmap representing the z-axis displacement of each taxel (as shown in Fig. \ref{fig:video_input}) to create a single video stream. The cmthermal color map \cite{cmthermal} is applied to the heatmap to make visual changes in tactile intensity intuitively recognizable.

The second format is a time-series representation designed for the GSP 
to capture mechanical state transitions. 
A data cleaning process is applied to the tactile sensor data. 
For both the left and right tactile sensors, the following steps are performed:
\begin{itemize}
    \item Clipping negative values to zero.
    \item Computing the average z-axis displacement across all 16 taxels.
    \item Normalizing the values to the range [0,\ 1].
\end{itemize}
At each time step, a three-dimensional feature vector, which corresponds to the Sensor Vector $\bm{u}_t$ in Fig. \ref{fig:propose_method}, is constructed by concatenating the cleaned tactile values and the grasping force, resulting in Eq. \ref{eq:vec}. 
\begin{equation}
\label{eq:vec}
\bm{u}_t =
\begin{bmatrix}
U_t^{\text{tactile left}} , \
U_t^{\text{tactile right}} , \
U_t^{\text{grasping force}}
\end{bmatrix}
\in \mathbb{R}^3
\end{equation}
During grasping, the applied force is increased from 0.1 N to 1.4 N in 
increments of 0.1 N, yielding 14 discrete time steps. By stacking these 
feature vectors over time, we obtain a time-series matrix 
$\bm{u}_{t} \in \mathbb{R}^{3 \times 14}$ that represents the evolution of 
mechanical interactions during grasping.

\subsection{Supervisor Generator (SG)}

We propose the SG to eliminate the prohibitive cost of manual annotation for newly encountered objects by establishing a few-shot subjective grounding mechanism. VLMs frequently suffer from hallucinations and unstable boundary judgments when analyzing continuous physical transitions, such as the exact onset of object deformation. Because the perception of an excessive state is inherently subjective and object-dependent, generic linguistic instructions alone are insufficient to ensure the high-precision labeling required for robotic control.

To overcome this, the SG utilizes a minimal set of human-annotated trials as contextual anchors. 
Specifically, the VLM receives a text prompt and an unannotated target video to be labeled, together with a set of reference videos paired with their human-annotated timestamp labels, which serve as few-shot examples.
By treating these pairs as explicit baselines, the model calibrates its internal judgment to the human criteria. This strategy enables the VLM to propagate subjective expectations across the remaining unannotated trials of the same object category. 


To facilitate seamless integration between the SG and the GSP, the output is confined to a predefined JSON format. This enables the automated parsing of timestamps and reasoning strings, creating a robust data pipeline.

Ultimately, the SG serves as a vital bridge that transforms minimal human intervention into an object-wise training dataset. By aligning the VLM assessment with specific human criteria through few-shot anchoring, the framework enables the robot to learn human-centric control with significantly reduced manual labor.

%
%
%


\subsection{Grasp State Predictor (GSP)} \label{sec:gsp}
We propose the GSP as a lightweight module that performs sequential inference to predict the grasp state at the one-step-ahead (t+1) (Fig. 2, lower-right). The GSP transfers qualitative human visual criteria into quantitative robot execution by associating subjective state labels with the corresponding mechanical signatures of different materials.



At each time step $t$, the system receives a sensor vector $\bm{u}_t$ comprising the left and right tactile values and the gripper force to anticipate upcoming state transitions. To implement this real-time predictive capability, we employ an echo state network (ESN) \cite{jaeger2001} consisting of an input layer, a high-dimensional reservoir layer, and a linear readout layer. The reservoir layer acts as a non-linear temporal filter that captures the complex dynamics and history of the physical interaction signals. Because the internal reservoir weights remain fixed, the network functions as a high-dimensional kernel that maps temporal patterns into a space where they become linearly separable. This architecture significantly reduces computational costs by restricting optimization solely to the final readout layer via ridge regression, which provides an efficient closed-form solution. Rather than maintaining a single set of universal parameters, our system preserves specific readout weight matrices individually for each newly encountered object category. 
By simply switching the weight matrix, the system achieves instantaneous adaptation to unique deformation characteristics. 



\section{EXPERIMENTS}
To verify the effectiveness of the proposed framework in bridging the semantic gap and achieving rapid adaptation, we conduct evaluations structured as follows:

\begin{itemize} 
    \item A. Dataset and Setup: Details of multimodal data collection and ground truth definitions for three distinct objects to establish the experimental baseline. 
    \item B. SG Reliability: Evaluation of label accuracy versus example count to demonstrate the mitigation of manual data collection bottlenecks. 
    \item C. GSP Performance: Validation of the lightweight prediction model through four specific criteria: 
    \begin{enumerate} 
        \item Adaptability: Measurement of convergence speed during incremental training to verify the capability for rapid online adaptation to newly encountered objects. 
        \item Predictability: Assessment of one-step-ahead state estimation accuracy to ensure the fundamental predictive performance of the model. 
        \item Real-time Stability: Verification of system performance during physical robot transport tasks to ensure operational reliability in dynamic environments. 
        \item Human Subjective Alignment: Comparison of robotic behavior with assessments from 25 participants to validate the bridging of the semantic gap. 
    \end{enumerate} 
\end{itemize}



\subsection{Experimental Setup and Dataset Collection}

To evaluate the proposed framework, a specific experimental dataset was constructed according to the data preparation protocol established in Section \ref{sec:dataset_preparation}. For hardware setup, data collection utilizes the gripper of a Toyota Human Support Robot \cite{yamamoto2019}, as illustrated in the Robot System and Preprocessing section of Fig. \ref{fig:propose_method}. To capture the multimodal data, a TierIV C1 image sensor \cite{tier4_c1} mounted on top of the gripper records RGB images. Furthermore, XELA uSkin tactile sensors \cite{xela} attached to both fingertips of the gripper measure the displacement along the z-axis, which corresponds to the push direction. The system also synchronously records the applied grasping force of the robot.

\begin{figure}[t]
  \centering
  \includegraphics[width=0.55\columnwidth]{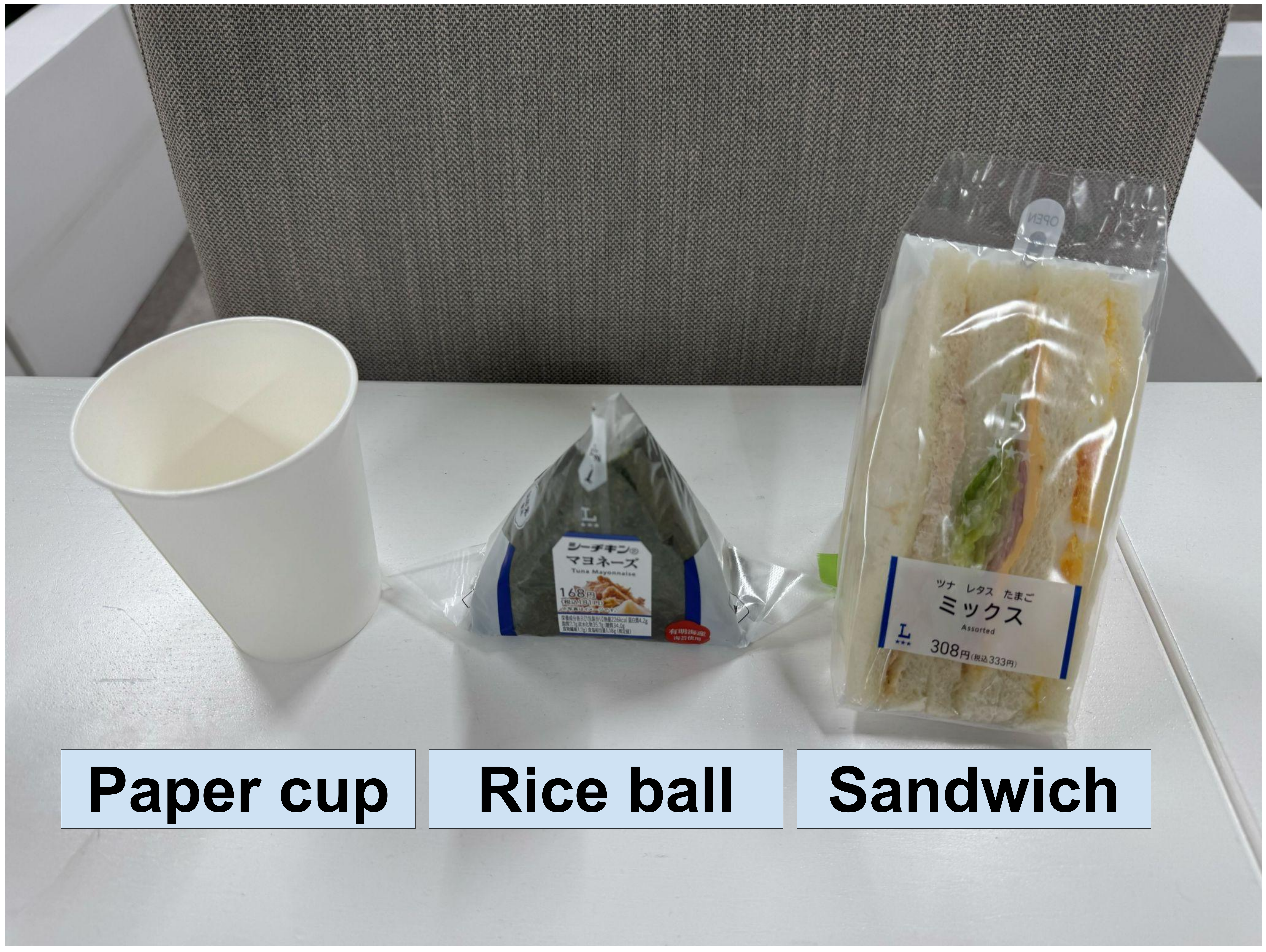}
  \caption{Target objects.}
  \label{fig:target_obj}
\end{figure}

As shown in Fig. \ref{fig:target_obj}, the target items include a rice ball, a sandwich, and a paper cup. These items were selected as representative deformable objects commonly found in unstructured environments such as homes and convenience stores, guided by the official rulebook of the World Robot Summit Future Convenience Store Challenge \cite{wrs2025_fcsc_rulebook}.

During the data collection phase, the robot executed four independent grasping trials for each of the three target objects, yielding a total of 12 data sequences. In each trial, the robot incrementally applied grasping force from 0.1 N to 1.4 N to capture the multimodal sensor data. To establish the absolute GT for these experiments, a human annotator assigned the three discrete state labels to all 12 acquired sequences utilizing the visual deformation criteria defined in the methodology. This comprehensively annotated dataset serves as the definitive baseline to evaluate the inference accuracy of the SG in Section \ref{sec:eval_sg}, as well as the adaptability and predictability of the GSP in Section \ref{sec:eval_gsp}.

\subsection{Evaluation of Supervisor Generator} \label{sec:eval_sg}
To evaluate the accuracy of the SG in assigning timestamps to unannotated data, the experimental setup employs the Google Gemini 3 Pro API \cite{gemini} as the core VLM. Because the proposed framework relies on few-shot propagation, this evaluation specifically investigates the impact of the amount of human-provided context on estimation accuracy. The experiment compares three distinct prompt configurations by providing zero, one, and two annotated example videos per object category. Consequently, as the number of provided examples increases, the number of remaining unannotated target sequences evaluated by the model naturally decreases from the original 12 trials to 9 and 6, respectively. 
To ensure robust evaluation and avoid bias in example selection, a 4-fold cross-validation scheme is employed, where the trials are partitioned into four subsets and each subset is alternately used for evaluation while the remaining subsets provide annotated examples according to the prompt configuration. This comparison explicitly demonstrates how effectively human subjective criteria transfer to new trials based on the volume of initial guidance.

\begin{table*}[t]
\centering
\caption{Accuracy and F1 Scores of the SG Across Experimental Settings for Each Object.}
\label{tab:sg_f1_all}
\begin{tabular}{lcccccc}
\toprule
\multirow{2}{*}{Objects}
& \multicolumn{2}{c}{Zero annotated examples}
& \multicolumn{2}{c}{One annotated example}
& \multicolumn{2}{c}{Two annotated examples} \\
\cmidrule(lr){2-3}\cmidrule(lr){4-5}\cmidrule(lr){6-7}
& Accuracy & F1 Score & Accuracy & F1 Score & Accuracy & F1 Score \\
\midrule
rice ball        & 0.74 & 0.64 & 0.94 & 0.92 & 1.00 & 1.00 \\
sandwich  & 0.84 & 0.71 & 1.00 & 1.00 & 1.00 & 1.00 \\
paper cup        & 0.72 & 0.67 & 0.97 & 0.95 & 0.97 & 0.94 \\
\bottomrule
\end{tabular}
\end{table*}

Table \ref{tab:sg_f1_all} presents the quantitative classification performance of the SG across the three prompt configurations. When provided with zero annotated examples, the model yielded the lowest accuracy and F1 scores across all target items. This baseline performance highlights the inherent difficulty the VLM faces in autonomously establishing appropriate grasping criteria without human guidance.
Providing a single example video significantly improved the overall scores. However, detailed analysis revealed that the model still occasionally misclassified adjacent boundary frames, specifically confusing the sliding state with the appropriate state, and the appropriate state with the excessive state. 
Conversely, providing two annotated examples effectively resolved these boundary ambiguities. Under this configuration, the SG achieved higher classification accuracy and F1 scores for the rice ball and the sandwich, while maintaining a high accuracy of 0.97 and an F1 score of 0.94 for the paper cup. 

These results indicate that propagating merely two human-annotated trials provides sufficient contextual grounding for the model to reliably transfer subjective human visual criteria to unannotated data, thereby enabling reliable training data generation for the subsequent prediction module, as reflected in the improved classification performance.

\subsection{Evaluation of Grasp State Predictor} \label{sec:eval_gsp}
The GSP was evaluated from four perspectives: adaptability (\ref{sec:gsp_adaptability}), predictability (\ref{sec:gsp_predict}), stability (\ref{sec:gsp_stability}), and agreement with human subjective evaluations (\ref{sec:human_subjective}). 
This evaluation isolates the predictive capability of the GSP from possible timestamp estimation errors introduced by the SG. Therefore, the reported GSP scores represent the upper-bound performance of the prediction module under human-annotated labels, rather than the full end-to-end SG-to-GSP pipeline.

We use Optuna to optimize the hyperparameters and select the combination that minimizes the root mean square error (RMSE) for waveform tracking on the test data.
We fix this hyperparameter set across all four GSP experiments to eliminate tuning bias and ensure a fair comparison.
Table~\ref{tab:hp_reservoir_v2} lists the hyperparameters of the ESN and ridge regression used in the experiments.
\begin{table}[tb]
  \centering
  \caption{Hyperparameters of the GSP used in the experiments}
  \label{tab:hp_reservoir_v2}
  \begin{tabular}{lr}
    \toprule
    Parameter & Value \\
    \midrule
    Number of ESN nodes & 628 \\
    Input scaling factor & 0.671 \\
    Recurrent scaling factor & 0.366 \\
    Sparsity of connections & 0.816 \\
    Leak rate & 0.808 \\
    Regularization coefficient of ridge regression & 0.0045 \\
    \bottomrule
  \end{tabular}
\end{table}

\subsubsection{Evaluation of Adaptability} \label{sec:gsp_adaptability}

To evaluate the online adaptability of the GSP to a newly encountered object, the experiment investigates the stabilization speed of the prediction model during continuous operation. As detailed in \ref{sec:gsp}, the system incrementally updates object-specific readout weights. To practically evaluate this capability, the experiment sequentially expands the training data from one to three sequences for each target object. At each incremental step, the newly updated weights are evaluated against the remaining unseen trials of that specific object. This progressive evaluation explicitly demonstrates how rapidly the state predictions converge to the established human-centric criteria with minimal physical interactions.

\begin{figure}[t]
  \centering
  \includegraphics[width=0.8\columnwidth]{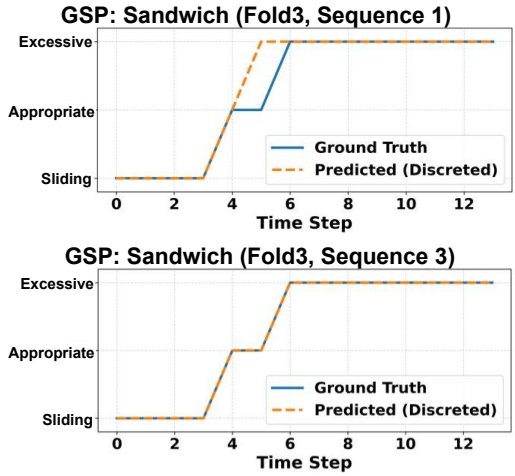}
  \caption{Adaptability of the GSP: Comparison of prediction results when trained on one versus three data sequences.}
  \label{fig:gsp_offline}
\end{figure}

Figure \ref{fig:gsp_offline} compares the prediction results for the sandwich object when trained with one and three sequences.
The vertical axis of the graph represents the three discrete states of sliding, appropriate, and excessive, while the horizontal axis indicates the time-series steps as the grasping force steps. The solid blue line represents the GT, and the dashed orange line shows the prediction results of GSP; a closer match between the two indicates a more successful prediction. As shown in Fig. \ref{fig:gsp_offline}, the boundaries of the state transitions remain unstable in the initial stage with only one training sequence. However, by increasing the training data to three sequences, the prediction results successfully matched the GT.

These results suggest that the GSP can rapidly adapt to the new evaluation criteria of newly encountered objects, even with a minimal amount of training data.

\subsubsection{Evaluation of Predictability} \label{sec:gsp_predict}
To assess the predictability of the fully adapted GSP, we quantitatively evaluated whether it can accurately classify the grasp state one-step-ahead using the sensor information $\bm{u}_{t}$. For each object, a 4-fold cross-validation was conducted, allocating three data sequences for training and the remaining one for testing. In addition to Accuracy and F1 scores, the IoU@Appropriate score was employed as an evaluation metric, which indicates how accurately the prediction results covered the appropriate state intervals defined by the GT.

Equation \ref{eq:iou05} defines the IoU@Appropriate metric. Here, $T_{GT}$ represents the set of time steps where the GT indicates the appropriate state, and $T_{pred}$ represents the set of time steps where the model predicts the appropriate state. The numerator, $|T_{GT} \cap T_{pred}|$, represents the duration where both the GT and the prediction agree on the appropriate state. The denominator, $|T_{GT} \cup T_{pred}|$, represents the duration where at least one of them indicates the appropriate state.
\begin{equation}
    \text{IoU@Appropriate} = \frac{|T_{GT} \cap T_{pred}|}{|T_{GT} \cup T_{pred}|}
    \label{eq:iou05}
\end{equation}

\begin{table}[tb]
\centering
\caption{Predictability of the GSP: Accuracy, F1 Score, and IoU@Appropriate.}
\label{tab:gsp_offline_result}
\begin{tabular}{lcccc}
\toprule
Object & Accuracy & F1 Score & IoU@Appropriate \\
\midrule
rice ball   & 1.00 & 1.00 & 1.00 \\
sandwich  & 1.00 & 1.00 & 1.00 \\
paper cup & 1.00 & 1.00 & 1.00 \\
\bottomrule
\end{tabular}
\end{table}

Table \ref{tab:gsp_offline_result} summarizes the experimental results regarding GSP predictability. The GSP achieved a score of 1.0 across all evaluation metrics for all objects. These results indicate that, within the controlled object-wise adaptation setting used in this study, the GSP can reproduce the grasp-state boundaries defined by human annotations from tactile and grasping force signals. However, because each fold contains a limited number of sequences, these results should be interpreted as evidence of feasibility in the present task setting rather than as proof of broad generalization.


\subsubsection{Evaluation of Stability} \label{sec:gsp_stability}
Subsequently, the GSP was integrated into the real-time control of a physical robot to evaluate its stability. This experiment aims to verify whether the GSP can maintain the grasping force within the appropriate state on a physical robot.

For the experimental procedure, the open gripper first approached each object placed on a table, and the grasping force was incrementally increased by 0.1 N until the GSP predicted that the next force step would enter the excessive state. The robot then selected the preceding force level, which was predicted as appropriate, and used it for the transport task. To verify whether the robot could transport the object while maintaining an appropriate grasping force, the robot lifted the object, rotated 90 degrees, moved approximately 2 m, and placed it down. We conducted this procedure three times for each of the three object types, totaling nine trials. As in Section \ref{sec:gsp_predict}, Accuracy, F1 score, and IoU@Appropriate were used as quantitative evaluation metrics for inference performance.


\begin{table}[tb]
\centering
\caption{Quantitative Evaluation of GSP Stability on the Real Robot.}
\label{tab:gsp_online_result}
\begin{tabular}{lcccc}
\toprule
Object & Accuracy & F1 Score & IoU@Appropriate \\
\midrule
rice ball   & 0.93 & 0.90 & 0.78 \\
sandwich  & 1.00 & 1.00 & 1.00 \\
paper cup & 1.00 & 1.00 & 1.00 \\
\midrule
All objects & 0.98 & 0.97 & 0.93 \\
\bottomrule
\end{tabular}
\end{table}

Table \ref{tab:gsp_online_result} presents the quantitative evaluation results using the real robot. Although the GSP achieved an overall Accuracy of 0.98 across all objects, the score remained at 0.93 in the rice ball trials. Focusing on the IoU@Appropriate for the rice ball, a discrepancy was observed in the appropriate interval between the GT and prediction.

\begin{figure}[t]
  \centering
  \includegraphics[width=0.9\columnwidth]{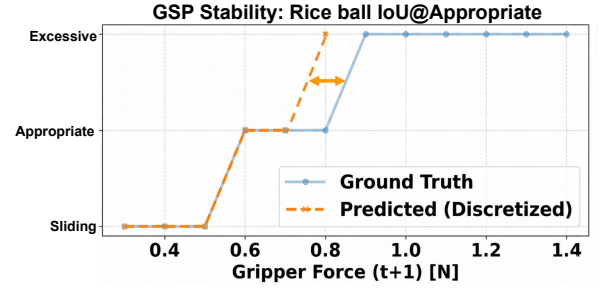}
  \caption{Stability of the GSP: Prediction Results of IoU@Appropriate for the rice ball.}
  \label{fig:rice ball05}
\end{figure}

As shown in Fig. \ref{fig:rice ball05}, in the rice ball trials, the prediction transitioned to the excessive state one time step earlier than the GT. 
This discrepancy originates from the asymmetrical geometry and non-uniform internal density of the rice ball, where even slight grasping deviations significantly alter the perceived mechanical resistance. Furthermore, its stiffness is inherently susceptible to environmental changes, such as temperature variations, between data collection and real-world execution. Because the model strictly acquires its criteria from limited uniform data, it conservatively interpreted this unlearned increase in physical resistance as reaching the deformation threshold. 

On the other hand, the actual motion stop timing based on prediction fell safely within the appropriate interval of the GT, and the robot successfully achieved the task objective without dropping or crushing the object. These physical robot evaluations demonstrate that utilizing the grasp state prediction of the GSP enables the system to maintain the grasping force in the appropriate state in real-time, ensuring stable transportation.

\subsubsection{Subjective Human Evaluation}\label{sec:human_subjective}
\begin{figure}[t]
   \centering
   \includegraphics[width=0.7\linewidth]{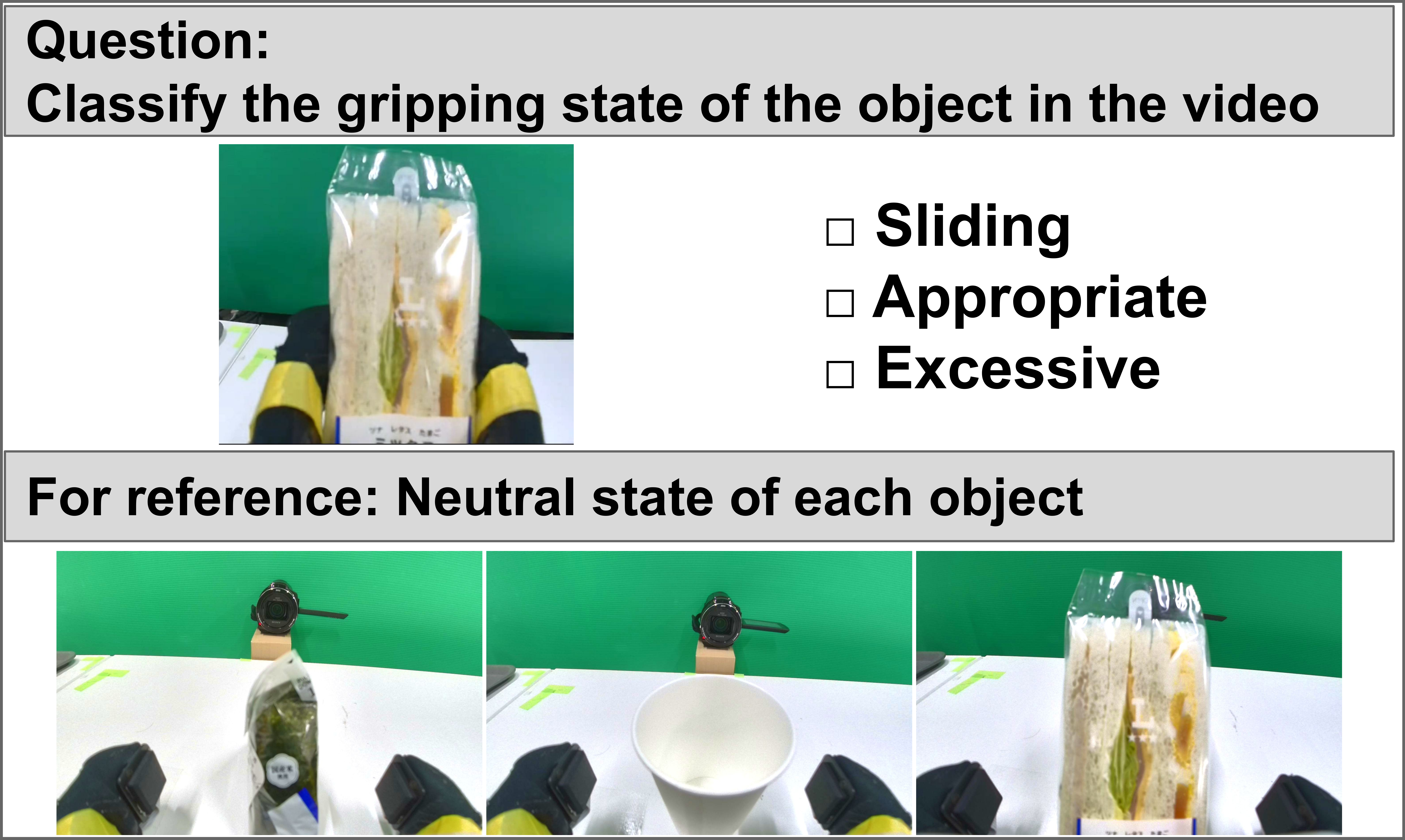}
   \caption{Example of the response section in the Google Forms questionnaire.}
   \label{fig:google_form}
\end{figure}

To verify whether the proposed framework bridged the gap between the robot behavior and human expectations of appropriateness, we conducted a subjective evaluation study with 25 participants. 
During this evaluation, the participants observed videos of nine distinct trials from the physical robot experiments detailed in Section \ref{sec:gsp_stability}, which included three sequences each for the rice ball (O1--O3), the sandwich (S1--S3), and the paper cup (P1--P3). For each recorded trial, the experimental procedure required the participants to intuitively classify the observed robotic grasp state into the sliding, appropriate, or excessive category. As illustrated in Fig. \ref{fig:google_form}, this subjective survey was administered through a Google Forms web-based questionnaire.


\begin{figure}[t]
   \centering
   \includegraphics[width=1.0\linewidth]{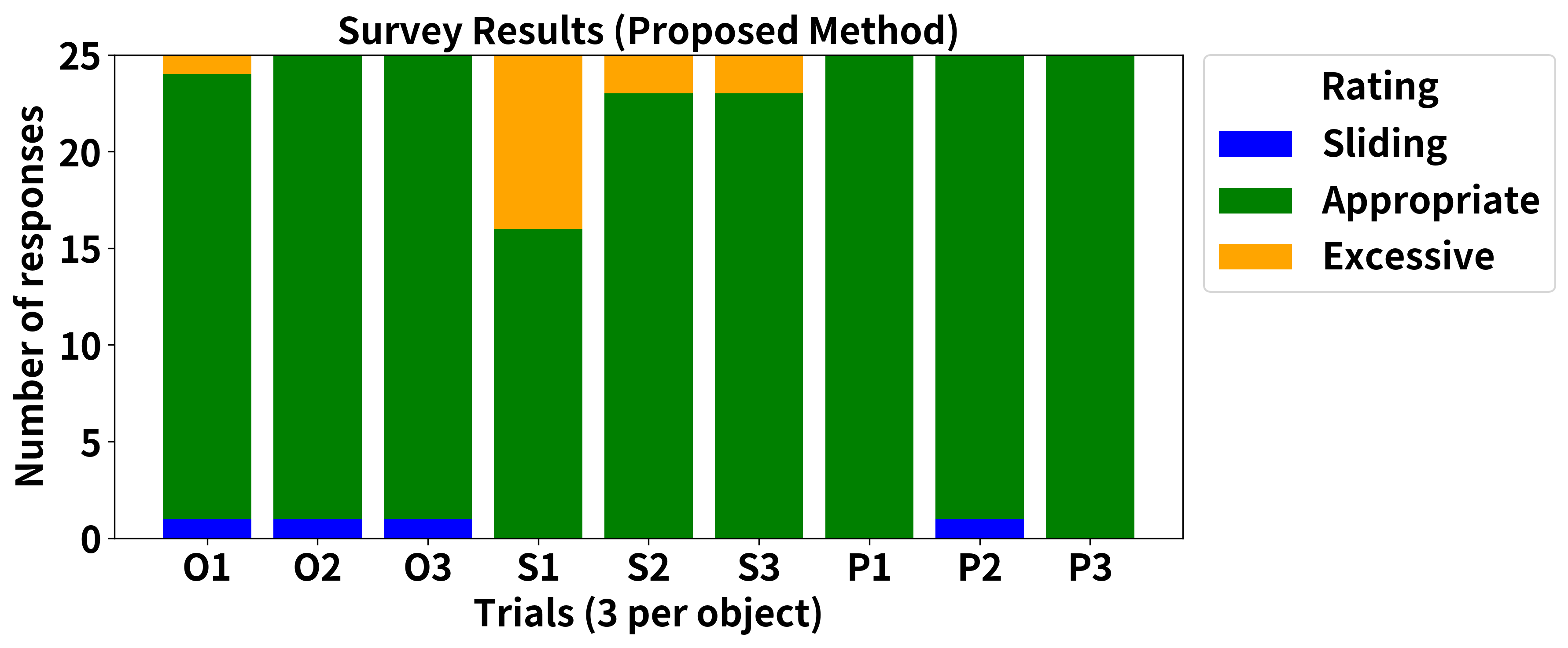}
   \caption{Results of the human subjective evaluation.}
   \label{fig:human_eval}
\end{figure}

Figure \ref{fig:human_eval} presents the results of this subjective evaluation. The vertical axis represents the number of respondents, while the horizontal axis indicates the nine physical robot trials. In the graph, blue indicates sliding, green indicates appropriate, and orange indicates excessive. 
A large number of responses evaluating the grasp as appropriate demonstrate the effectiveness of the grasping force adjustment by the GSP. Consequently, in all trials except S1, the proportion of participants who evaluated the grasping behavior of the robot as appropriate exceeded 90\%.

Regarding trial S1, where the evaluations were divided, we attribute this to an off-center grasping position that caused the object to appear slightly distorted visually. This visual distortion likely led some participants to judge the state as excessive.

\section{LIMITATIONS and Future Work}

The proposed framework has several limitations. First, the evaluation is restricted to three objects and a small number of sequences; therefore, the results should be interpreted as feasibility evidence rather than proof of broad generalization. The current framework may also require additional human annotations for object categories or intra-category variations with substantially different deformation and tactile-response characteristics, such as different types of sandwiches. To improve scalability, future work will use prediction errors and interactive human or language-based feedback as learning signals, and extend the evaluation to more diverse objects, users, and manipulation scenarios. Second, the current three-state representation is an initial approximation of human-centric grasp appropriateness, and future work should consider probabilistic or personalized preference models. Third, the GSP was evaluated using human-annotated labels to isolate its predictive capability, and full SG-to-GSP error propagation remains to be evaluated. Finally, future work will incorporate grasp-pose information, uncertainty-aware prediction, dynamic slip detection, and posture compensation to handle tactile-response variations.

\section{CONCLUSIONS}
This study presented a human-centric grasp state assessment framework that transfers visually perceived grasp appropriateness to tactile and grasping-force-based prediction for deformable objects. The results demonstrate the feasibility of VLM-assisted supervisor generation and lightweight ESN-based object-wise adaptation in the evaluated task setting. Future work will address full SG-to-GSP evaluation, broader object and task variation, and long-term adaptation using interactive human or language-based feedback.

\addtolength{\textheight}{-12cm}   


\section*{ACKNOWLEDGMENT}
    This paper is based on results obtained from project JPNP16007 commissioned by the New Energy and Industrial Technology Development Organization (NEDO). This work was supported by JSPS KAKENHI Grant Numbers 23H03468 and 24KJ182. This work was supported by JST ALCA-Next Grant Number JPMJAN23F3. This work was supported by JST SPRING, Japan Grant Number JPMJSP2154.


\bibliographystyle{ieeetr}
\bibliography{ref.bib}

\end{document}